\documentclass[10pt,journal,compsoc]{IEEEtran}

\ifCLASSOPTIONcompsoc
\usepackage[nocompress]{cite}
\else
\usepackage{cite}
\fi

\ifCLASSINFOpdf
\usepackage[pdftex]{graphicx}
\else
\usepackage[dvips]{graphicx}
\fi

\usepackage{amsmath}
\usepackage{amssymb}
\newcommand{\eg}{\textit{e}.\textit{g}.}

\begin{document}
\title{Interpretable Visual Question Answering by Reasoning on Dependency Trees}
\author{Qingxing Cao, Bailin Li, Xiaodan Liang and Liang Lin
	\IEEEcompsocitemizethanks{
		\IEEEcompsocthanksitem This work was supported in part by the National Key Research and Development Program of China under Grant No. 2016YFB1001004, in part by National Natural Science Foundation of China (NSFC) under Grant No. 61976233, 61836012, 61622214, and in part by the Natural Science Foundation of Guangdong Province under Grant No. 2017A030312006.
		\IEEEcompsocthanksitem Q.Cao and X. Liang are with the School of Intelligent Systems Engineering, Sun Yat-sen University, China. B. Li is with DMAI Great China. L. Lin is with the School of Data and Computer
		Science, Sun Yat-sen University, China
		\IEEEcompsocthanksitem Corresponding author:
		Xiaodan Liang (E-mail: xdliang328@gmail.com)}}
% The paper headers
\markboth{IEEE Transactions on Pattern Analysis and Machine Intelligence,~Vol.~X, No.~X, XXX}%
{Shell \MakeLowercase{\textit{et al.}}: Bare Demo of IEEEtran.cls for Computer Society Journals}
\IEEEtitleabstractindextext{
	\begin{abstract}
		Collaborative reasoning for understanding image-question pairs is a very critical but underexplored topic in interpretable visual question answering systems. Although very recent studies have attempted to use explicit compositional processes to assemble multiple subtasks embedded in questions, their models heavily rely on annotations or handcrafted rules to obtain valid reasoning processes, which leads to either heavy workloads or poor performance on compositional reasoning. In this paper, to better align image and language domains in diverse and unrestricted cases, we propose a novel neural network model that performs global reasoning on a dependency tree parsed from the question; thus, our model is called a parse-tree-guided reasoning network (PTGRN). This network consists of three collaborative modules: i) an attention module that exploits the local visual evidence of each word parsed from the question, ii) a gated residual composition module that composes the previously mined evidence, and iii) a parse-tree-guided propagation module that passes the mined evidence along the parse tree. Thus, PTGRN is capable of building an interpretable visual question answering (VQA) system that gradually derives image cues following question-driven parse-tree reasoning. Experiments on relational datasets demonstrate the superiority of PTGRN over current state-of-the-art VQA methods, and the visualization results highlight the explainable capability of our reasoning system.
	\end{abstract}
	\begin{IEEEkeywords}
		Visual Question Answering, Image and Language Parsing, Deep Reasoning, Attention Model
\end{IEEEkeywords}}
% make the title area
\maketitle
\IEEEdisplaynontitleabstractindextext
\IEEEpeerreviewmaketitle
\IEEEraisesectionheading{\section{Introduction}\label{sec:introduction}}
\IEEEPARstart{T}{he} task of visual question answering (VQA) is to predict the correct answer given an image and a textual question. 
The key to this task is the ability to apply coreasoning over the image and language domains.
However, most previous methods~\cite{Seq2Seq, HiCoAtt, MLB} work in a manner similar to a black box, i.e., simply mapping the visual content to the textual words by crafting neural networks. The main drawback of these methods is the lack of interpretability of the results, i.e., why are these answers produced?
Moreover, it has been shown that the accuracy of these results may be improved by overfitting the data bias in the VQA benchmark~\cite{balanced_vqa_v2} and that not explicitly exploiting the structures of the text and images leads to unsatisfactory performance on relational reasoning~\cite{clevr}.
Very recently, a few pioneering works~\cite{e2emn,inferring2017, GraphVQA} have taken advantage of the inherent structure of text and images; these works parse the question-image input into a tree or graph layout and assemble local features of nodes to predict the answer. For example, the layout ``\emph{more}(\emph{find}(\emph{ball}),\emph{find}(\emph{yellow}))'' means that the module should locate the ball and the yellow object in the image first and then combine the two results to determine whether there are more balls than yellow objects.
However, these methods rely on either handcrafted rules for understanding questions or a layout parser that is trained from scratch; the first approach requires human experts to design appropriate rules in a specific domain or requires heavy labor to annotate a specific dataset~\cite{NSVQA}, while the second results in a large decrease in performance~\cite{inferring2017}. We argue that these limitations severely limit the application potential of these approaches for understanding general image-question pairs that may contain diverse and open-ended question styles.
\begin{figure}[t]
	\includegraphics[width=0.5\textwidth]{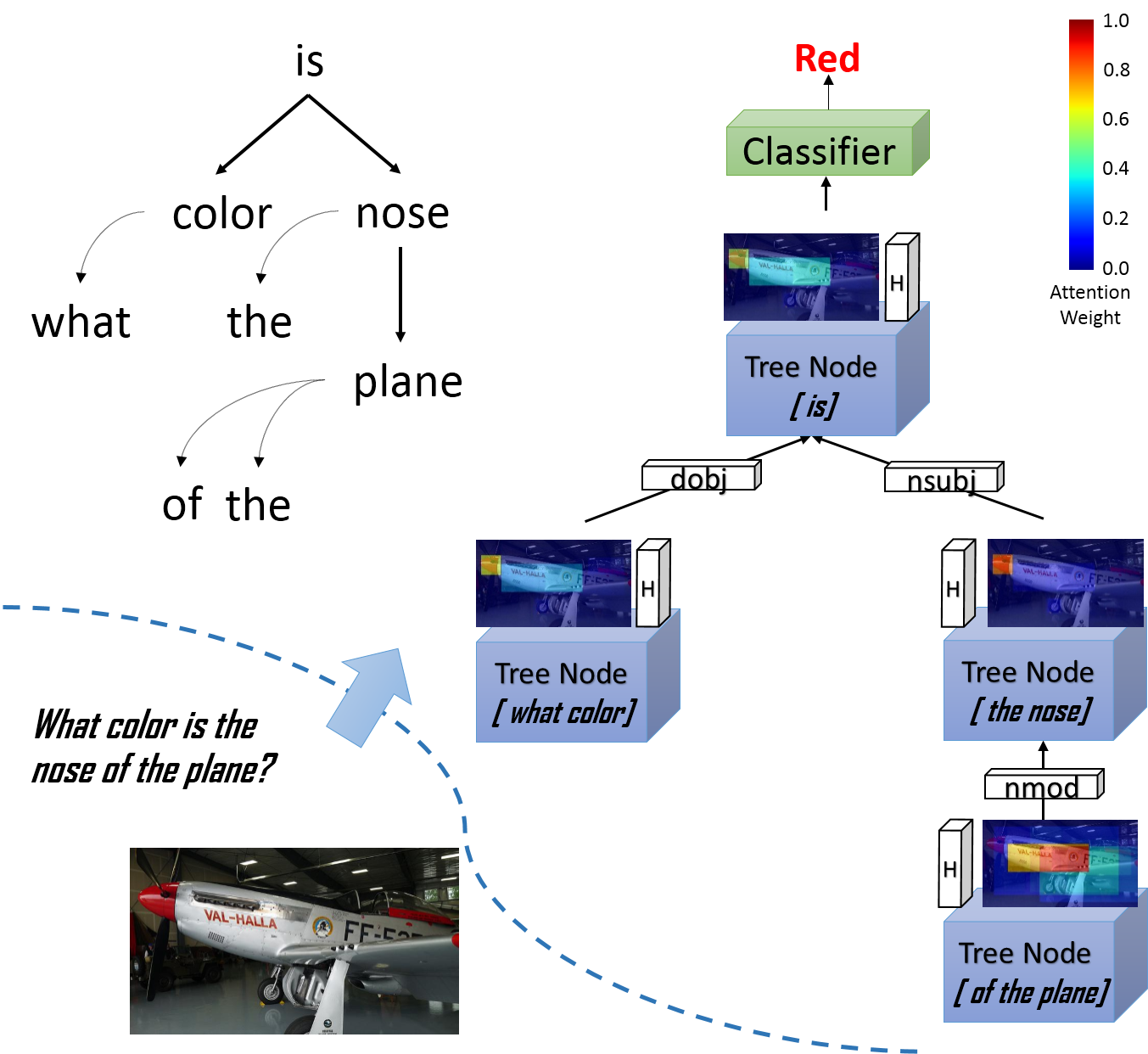}
	\caption{Illustration of our parse-tree-guided reasoning network (PTGRN) that sequentially performs reasoning over a dependency tree parsed from the question. Conditioned on preceding word nodes, PTGRN alternately mines visual evidence for nodes via an attention module and integrates the features of child nodes via a gated residual composition module.
	}
	\label{fig:intro}
\end{figure}

To achieve a general and powerful reasoning system that can enable reasoning over any dependency parse trees of questions without domain-specific knowledge, we propose a novel parse-tree-guided reasoning network (PTGRN) that contains three collaborative modules to perform tailored reasoning operations for addressing the two most common word relations in the questions. As shown in Figure~\ref{fig:intro}, given a specific dependency tree of a question parsed by an off-the-shelf parser, we construct a reasoning route that follows the parse tree layout, which is a tree structure composed of several types of nodes or edges. Our proposed network then alternately applies three collaborative modules on each word node for global reasoning to 1) exploit the local visual evidence of each word guided by the exploited regions of its child nodes, 2) integrate the messages of child nodes via gated residual composition, and 3) propagate the hidden and exploited map toward its parent with respect to the type of edges. Notably, in contrast to previous methods, PTGRN is a general and interpretable reasoning VQA framework that does not require any complicated handcrafted rules or ground-truth annotations to obtain a specific layout.

Specifically, we observe that the frequently used types of dependency relations can be categorized into two sets depending on whether the head is a predicate that describes the relation of its children (\eg \emph{color} $\leftarrow$ \emph{is}, \emph{is}$\rightarrow $\emph{nose}) or a word described by its child (\eg \emph{furthest}$\rightarrow$\emph{object}). We refer to the first set as a clausal predicate relation, which describes how a parent node composes its children, and we refer to the second as a modifier relation, which will help specify an object more concretely given the parent-child pairs.
Thus, we design an attention module to unitize the information propagated from the modifier relations, a gated residual composition module to compose the messages from child nodes, and finally, a parse-tree-guided propagation module to transfer the node's inner representations to its parent conditioned on the fine-grained relation types.

First, the attention module mines visual evidence from the image feature map given the word and encoded attention maps from child nodes. We sum the encoded attention maps from child nodes and fuse the result with the image feature and word encoding. Then, we perform an attention operation on the fused hidden map to extract new local visual evidence for the current node.
Second, two gated residual composition modules separately integrate both the mined local visual evidence and the attention map with the child nodes. To retain the information from an arbitrary number of child nodes, the module sums over the child nodes and learns a gate and a residual that will forget and update the hidden representations. Finally, the parse-tree-guided propagation module transforms the composed visual hidden representation and the attention hidden representation based on a head-dependent relation type and propagates the output message to the parent nodes.  This edge-dependent module is capable of learning to encode how much of a hidden vector should be persevered given a specific head-dependent type. Thus, the gated residual composition module of parent nodes can forget a previous message if necessary. The hidden and attention output message of the root node will pass through a multilayer perceptron to predict the final answer. 

A preliminary version of this work is published in CVPR2018. In this work, we inherit the idea of reasoning along the dependency parse tree, but we redevelop both modules such that they depend on specific types of relations rather than on a coarse-grained modifier relation and a clausal predicate relation. Furthermore, to improve the performance on question types of ``count'' and ``compare number'', we compose and propagate an attention map that is the same as the hidden representations. These changes involve fewer manually designed structures and thus lead to better performance and generalizability. We perform additional experiments to show the influence of redeveloped components, and we evaluate the generalizability against different tasks.

Extensive experiments show that our model can achieve state-of-the-art VQA performance on the CLEVR and FigureQA relational datasets. Moreover, the qualitative results further demonstrate the interpretability of PTGRN on collaborative reasoning over the image and language domains.

Our contributions are summarized as follows. 1) We present a general and interpretable reasoning VQA system that follows a general dependency layout composed of modifier relations and clausal predicate relations. 2) An attention module is proposed to enforce efficient visual evidence mining, a gated residual composition module is proposed for integrating knowledge of child nodes, and a parse-tree-guided propagation module is proposed to propagate knowledge along the dependency parse tree.

\begin{figure*}[t!]
	\centering
	\includegraphics[width=1.0\textwidth]{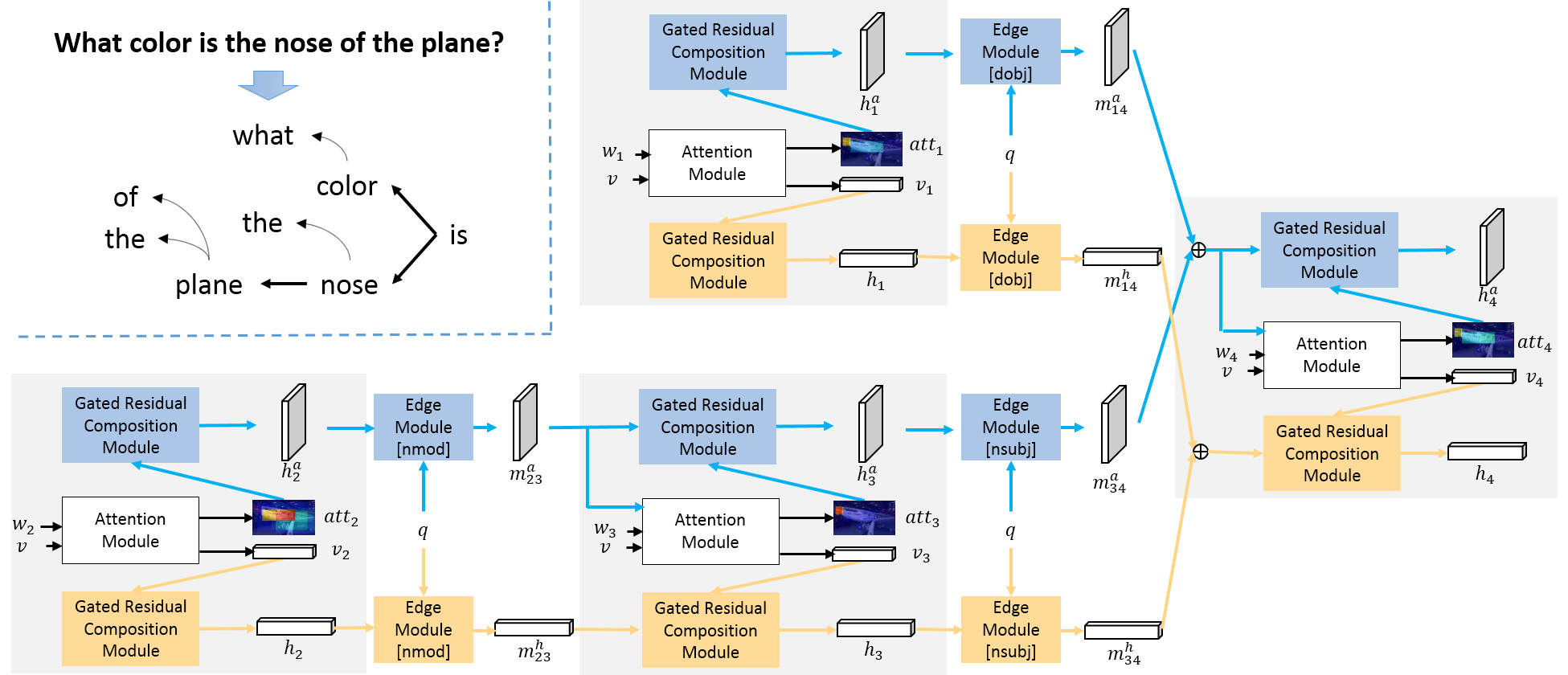}
	\hfill
	\caption{PTGRN pipeline. Each PTGRN module is composed of an attention module and two gated residual composition modules. Each node receives the encoded attention map, the hidden features from its children, and the image feature and word encoding. The attention module is employed to generate a new attention map conditioned on image features, word encodings and previous attended regions given by the child nodes. The gated residual composition module is trained to evolve a higher-level representation by dropping and integrating features of its children with local visual evidence. The edge modules transform the output attention and hidden feature according to the question encoding and the relation types (\emph{nmod: nominal modifier}, \emph{dobj: direct object} and \emph{nsubj: nominal subject}). The blue arrows indicate the propagation process of the attention map, and the yellow arrows represent the process of the visual hidden representation.}
	\label{fig:overview}
\end{figure*}

\section{Related Works}
\emph{Visual question answering.} \quad The VQA task requires coreasoning over both image and text to infer the correct answer.
The baseline method proposed in the VQA dataset~\cite{VQA} to solve this task uses a convolutional neural network (CNN)-long short-term memory (LSTM)-based architecture, which consists of a CNN to extract image features and an LSTM to encode the question features. The method combines these two features to predict the final answer. In recent years, a large number of works have followed this pipeline and have achieved substantial improvements over the baseline model. Among these works, the attention mechanism~\cite{DBLP:journals/corr/IlievskiYF16,Xu2016,shih2016att,Visual7W,StackedAtt,HiCoAtt} and the joint embedding of image and question representation~\cite{MCB,MLB,MFB} have been widely studied. The attention mechanism learns to focus on the most discriminative subregion rather than the whole image, providing a certain extent of reasoning to the answer. Different attention methods, such as stacked attention~\cite{StackedAtt} and coattention between question and image on different levels~\cite{HiCoAtt} consistently improve the performance of the VQA task. For multimodal joint embedding, Fukui \emph{et~al.}~\cite{MCB}, Kim \emph{et~al.}~\cite{MLB} and Hedi \emph{et~al.}~\cite{MUTAN} exploited the compact bilinear method to fuse the embedding of image and question and incorporate the attention mechanism to further improve performance.

However, some recently proposed works~\cite{balanced_vqa_v2,ECCV16baseline} have shown that the promising performance of these deep models might be achieved by exploiting the dataset bias. It is possible to perform equally well by memorizing the question-answer (QA) pairs or encoding the question with the bag-of-words method. To address this concern, newer datasets have been released recently. The VQAv2 dataset~\cite{balanced_vqa_v2} was proposed to eliminate data biases by balancing QA pairs. The CLEVR~\cite{clevr} dataset consists of synthetic images and provides more complex questions that involve multiple objects. This dataset has also balanced the answer distribution to suppress the data bias.

\emph{Reasoning model.} \quad Some prior works attempted to explicitly incorporate the knowledge into the network structure. \cite{externVQA, FVQA} encoded both images and questions into discrete vectors, such as image attributes or database queries. These vectors enable the model to query external data sources for common knowledge and basic factual knowledge to answer questions. \cite{zhu2017cvpr} actively acquired predefined types of evidence to obtain external information and predict answers. Other recent works proposed networks to handle compositional reasoning. \cite{DynaMem} augmented a differentiable memory and encoded long-term knowledge to infer answers. The neural reasoning network was recently proposed to address compositional visual reasoning. Rather than using a fixed structure to predict the answer to every question, this line of work assembles a structure layout for different questions into predefined subtasks.
Then, a set of neural modules is designed to solve a particular subtask.
Some representative works~\cite{e2emn,inferring2017} used the sequence-to-sequence recurrent neural network (RNN) to predict the postorder of the layout and jointly trained the RNN and the neural module on the CLEVR dataset by using reinforcement learning or expectation maximization. However, RNN training requires ground-truth layouts for supervision. The performance drops rapidly on CLEVR if ground-truth layouts are not used. 
On the VQA dataset, \cite{nmn,lnmn,e2emn} generated layouts based on dependency parsing. These three works first filter the set of dependencies to those connected to the wh-word at a certain distance. Then, they applied predefined modules to the remaining words to generate their inference layout. \cite{e2emn} further used the generated layouts as supervision to train an RNN layout parser. The parse tree filter and module assignment include too many manually designed rules and thus it is difficult to generalize across tasks. Our work intends to address this problem by applying a modular network to the raw parse tree.  
More recently, \cite{TbD} proposed an explicitly interpretable modular network by restricting the modules to pass only attention masks. \cite{NSVQA} converted both question and images into symbolic representations and executed symbolic program for reasoning. These works rely on ground-truth layouts in CLEVR but achieve high accuracy and interpretable results. 
In addition to the tree-structured layout, \cite{mac,SNN} performed sequential reasoning with augmented memory. At each step, a time-dependent module reads the memory based on both the question and the image and writes a newly generated encoding to the memory for answer prediction.
Different from these works, our model intends to achieve similar interpretable compositional reasoning on tree structures~\cite{e2emn,inferring2017,nmn,lnmn} without ground-truth layouts or manually designed rules~\cite{mac,SNN}.

\emph{Reasoning with counting-like questions.} \quad Counting the objects in an image has been studied for years. Lempitsky and Zisserman~\cite{VScount} learned to produce a density map for counting. ~\cite{DBLP:conf/cvpr/ZhangLWY15, DBLP:conf/eccv/Onoro-RubioL16} exploited a convolution network to estimate density and count target objects.
Zhang \emph{et~al.}~\cite{DBLP:journals/ijcv/ZhangMSSBLSPM17} studied the problem of salient object subitizing, which recognizes the number of salient objects when only a few objects are present. 
Chattopadhyay \emph{et~al.}~\cite{DBLP:conf/cvpr/ChattopadhyayVS17} further extended this approach by employing a divide and conquer strategy; they divided the image into subregions, counted within the regions individually, and combined the results.

However, the counting problem has been infrequently addressed in VQA. Anderson \emph{et~al.}~\cite{Anderson2017up-down} improved the ``counting'' question accuracy on VQAv2~\cite{balanced_vqa_v2} by first training an object detector on the Visual Genome dataset~\cite{vg}. The pretrained detector was used to extract the objects from the VQAv2 dataset, and the attention operation was performed on candidate objects rather than the grid feature map only. 
Trott \emph{et~al.}~\cite{count} followed this pipeline, representing each image as a set of detected objects. Then, they treated the counting problem as a sequential decision process by deciding which candidate object should be counted at each step. Finally, they employed reinforcement learning to train the model. However, this pipeline is limited to counting questions and cannot be easily applied to other types of questions.
Zhang \emph{et~al.}~\cite{learn2count} reported that it is the soft attention mechanism that limits the counting capability of VQA models. If one should answer how many cats are in an image, then cats counted by the normalized attention map will have weights that sum to $1$. Weighted sum pooling on these cats results in a feature vector that is similar to a single cat. To resolve this problem, Zhang \emph{et~al.}~\cite{learn2count} first generated an attention weight for each detected object; then, they performed ``soft'' nonmaximum suppression on the weights to count the attended objects.

Inspired by these observations, we separately encode and propagate the mined visual evidence and attention map. The attention map at each node will be composed with its children's output through a convolution GRU and will be propagated to its parent. We preserve the feature of the attended region and utilize it to predict the counting questions.

\section{Tree-Structured Composition Reasoning Network}
\subsection{Overview}
Given free-form questions $Q$ and images $I$, our proposed PTGRN model learns to predict the answers $y$ and their corresponding explainable attention maps. We first generate the tree-structured layout by parsing the input question $Q$ into a parse tree with an off-the-shelf universal Stanford Parser~\cite{dep}. We prune leaf nodes that are not nouns to reduce computational complexity. 

We denote the tree-structured layout as a 3-tuple $G=(u,X,E)$, and $u=(v,q)$ represents the global attributes that contain the image feature $v$ and the question encoding $q$.  $X=\{w_i\}_{i=1:N}$ represents the nodes in the parsed tree, and $N$ is the number of nodes. Each node is associated with word encoding $w_i$ in the origin question $Q$. $E=\{e_{i,j}\}_{i,j=1:N}$ represents the set of edges in the parsed tree, and $e_{i,j}$ denotes the edge type between head node $j$ and its dependent node $i$ in the dependency parse tree.

We then perform bottom-up visual question reasoning on the parse tree. Specifically, the image feature $v$ is extracted from each image via any CNN pretrained on ImageNet (\eg, conv5 features from ResNet-152~\cite{DBLP:conf/cvpr/HeZRS16} or conv4 features from ResNet-101~\cite{DBLP:conf/cvpr/HeZRS16}). The word encoding $w$ is obtained with Bi-GRU~\cite{gru}. Each word in the question is first embedded as a $200$-dimensional vector, and then the words are fed into a bidirectional GRU. The final word encoding $w$ is the hidden vector of Bi-GRU at its corresponding position, and the hidden vector at the end of the question is extracted as question encoding $q$.

A node $j$ has several inputs: the image feature $v$, the question encoding $q$, the word encoding $w_j$, and the messages $\{m_{ij}\}$ from its input edge $\{e_{ij}\}$. The message $m_{ij} = [m^{a}_{ij},m^{h}_{ij}]$ includes the message $m^{a}_{ij}$ for the attention map and $m^{h}_{ij}$ for hidden representation. It generates the attention map $att_{j}$ with the attention module $f_a$ as well as the hidden representation $h_{j}$ and the attention map encoding $h^{att}_j$ with the gated residual composition module $f_h$.

The inputs of an edge $jk$ are the hidden representation $h_{j}$ and the attention map encoding $h^{att}_j$ generated by node $j$. Given the inputs and the edge type $e_{jk}$, edge $jk$ generates the message $[m^{a}_{jk},m^{h}_{jk}]$ with the parse-tree-guided propagation module $f_e$ and transfers these messages to node $k$.

We update each node from bottom to top by postorder traversing the tree. Finally, at root node $N$, $[m^h_{N},m^a_{N}]$ are passed to a multilayer perceptron classifier to predict the answer, as shown in Figure~\ref{fig:overview}.

\begin{figure}[t!]
	\centering
	\includegraphics[width=0.45\textwidth]{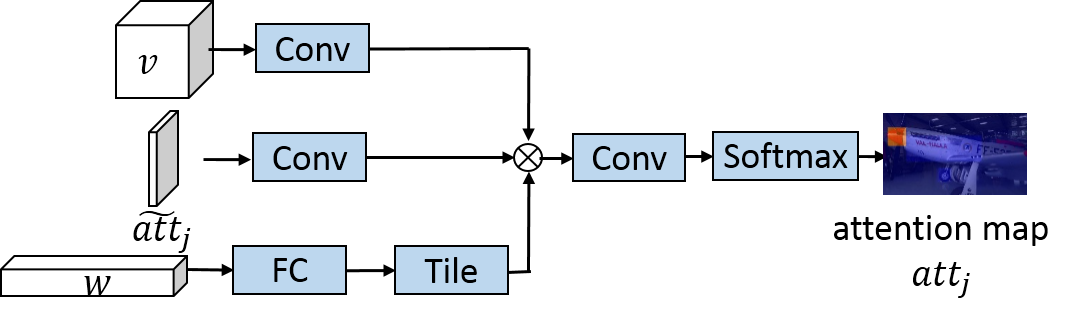}
	\caption{Detailed architecture of the attention module. The image feature, previously attended regions and word encoding are projected to $2048$-d features. Then, they are fused by elementwise multiplication. Finally, the fused feature is projected to a $1$-d attention map and normalized with softmax. }
	\label{fig:att}
\end{figure}
\begin{figure}[t!]
	\centering
	\includegraphics[width=0.45\textwidth]{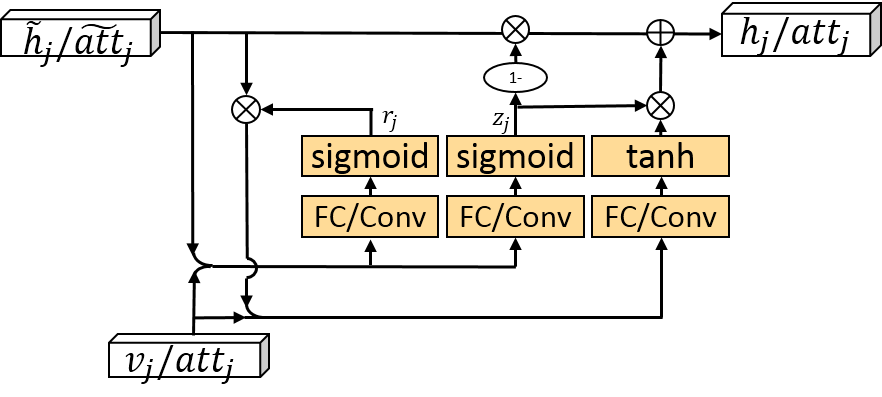}
	\caption{The gated residual composition module utilizes the architecture of a gated recurrent unit to integrate the features of its children with local visual evidence or attention map. The sum of children input is considered memory, and the local visual evidence or attention map is the input at the current step.}
	\label{fig:h}
\end{figure}
\begin{figure}[t!]
	\centering
	\includegraphics[width=0.45\textwidth]{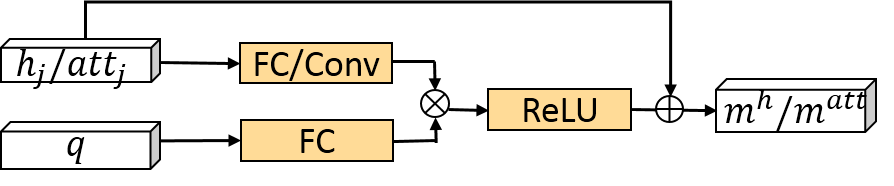}
	\caption{The parse-tree-guided propagation module performs bilinear fusion between the hidden/attention map and question encoding. Different edge types have the same architecture but different sets of weights.}
	\label{fig:edge}
\end{figure}

\subsection{Attention module} \label{attention_module}
In node $j$, the attention module $f_a$ is used to locate the image region that corresponds to the words encoding $w_j$ and the input attention message $\{m^a_{ij}\}_{e_{ij} \in E}$.
Specifically, as shown in Figure~\ref{fig:att}, the input attention feature $\tilde{att}_j$ of each node $j$ is first obtained by summing $\{m^a_{ij}\}_{e_{ij} \in E}$ as $\tilde{att}_j$.
Then, we project the image feature $v$, the input attention feature $\tilde{att}_j$ and the word encoding $w_j$ to $2048$-d features and perform elementwise multiplication on these three $2048$-d features.
Finally, the fused $2048$-d feature is fed into another convolution layer, resulting in a new attention map ${att}_{j}$. We further apply softmax to regularize the resulting attention map into the range $[0,1]$. The local image feature $v_j$ of node $j$ is then generated by the weighted sum of each grid in image feature $v$ given the weights in attention map ${att}_{j}$.

\subsection{Gated residual composition module} \label{GRCM}
A node $j$ contains two gated residual composition modules with similar architectures but different weights, as shown in Figure~\ref{fig:h}. This module is used to compose the newly generated attention map ${att}_{j}$ and local image feature $v_j$ with input messages $\{m^a_{ij}\}$ and $\{m^h_{ij}\}$ respectively.
In the preliminary work, the nodes share only weight at the same level, which contains many parameters and is not suitable for global reasoning with different parse trees.
However, applying a single node module multiple times essentially acts as a recurrent network and will suffer from gradient exploding/vanishing problems. 
Meanwhile, in the CLEVR dataset, there are a large number of objects referred to by their relationships with other objects, such as ``left of the big sphere'' or ``left of the brown metal''. To answer the question, the visual representation of ``big sphere'' is not necessary and might impact the final prediction. Previously, we dropped the hidden feature of these nodes when its head-dependent relation is a modifier relation. We want to enable the module to learn this drop process rather than using handcrafted rules. Here, we utilize the widely used gated recurrent unit to enable the module to learn the drop process.

As shown in Figure~\ref{fig:h}, the gated residual composition module $f_h$ first sums the messages of hidden representation $\{m^h_{ij}\}$ of its children into $\tilde{h}_j$, and then it concatenates $\tilde{h}_j$ with extracted local image feature $v_j$. Similar to a gated recurrent unit, the module generates the reset gate $r_j$ and the update gate $z_j$ based on the concatenated $[\tilde{h}_j, v_j]$. Then, the module produces an update vector $c_j$, updates and outputs the hidden representation $h_j$:
\begin{equation}
\begin{aligned}
\tilde{h}_j &= \sum_{(i,j) \in E}m^h_{ij} \\
z_j &= \sigma(W_z \cdot [\tilde{h}_j, v_j]) \\
r_j &= \sigma(W_r \cdot [\tilde{h}_j, v_j]) \\
c_j &= tanh(W \cdot [r * \tilde{h}_j, v_j]) \\
h_j &= (1-z_j)*\tilde{h}_j + z_j*c_j
\end{aligned}
\end{equation}
To encode the attention map $att_j$, we perform similar operations. We first sum $\{m^a_{ij}\}$ to obtain $\tilde{att}_j$. Since the attention map is a $2$-dimensional grid, we use the convolution gated recurrent unit to update the attention map encoding $h^{att}_j$. 
Compared with our preliminary work, which has $c_j = tanh(W \cdot [\tilde{h}_j, v_j])$ and $h_j = \tilde{h}_j + c_j$, in our current work, we add an update gate and a reset gate to the residual composition. These two gates can reduce the effect of the gradient exploding/vanishing problems when applying a single node multiple times, which enables us to reuse the node module across multiple parse tree heights.
The update process is also similar to that in Child-Sum Tree-LSTMs~\cite{treeLSTM}. In~\cite{treeLSTM}, the LSTM cell calculates forget gates for each of the memories from its child; we simplify this process by first summing over the children's hidden representation and outputting a single reset and forget gate for updating the hidden representation $h_j$.

\subsection{Parse-tree-guided propagation module}
Given the edge $jk$, the parse-tree-guided propagation module is used to transfer the hidden representation $h_j$ and the attention map encoding $h^{att}_j$ of node $j$ to messages $m^h_{jk}$ and $m^a_{jk}$ based on the edge type $e_{jk}$, as shown in Figure~\ref{fig:edge}.
This module to transfer the hidden representation to its parent node is based on the fine-grained dependency relation rather than directly blocking the hidden representation or the attention map based on two coarse-grained categories, namely, modifier relations and clausal predicate relations. For different types of edges, we apply the same module on hidden representation $h_j$ but with different sets of weights. The module performs multimodal bilinear pooling~\cite{MLB} on the hidden representation $h_j$ and the question encoding $q$. Then, it generates the message for its parent $m^h_{jk}$. 
Specifically, fully connected layers are applied to project both the hidden representation $h_j$ and the question encoding $q$ into two feature vectors that have the same size as $h_j$. Then, we perform elementwise multiplication on these two features and apply ReLU nonlinearity on the result. Finally, we add the result to $h_j$, resulting in a hidden vector $m^h_{jk}$ that will be passed through the edge.
\begin{equation}
m^h_{jk} = h_j + ReLU((W^h_{e_{jk}} \cdot h_j + b^h_{e_{jk}}) * (W^q_{e_{jk}} \cdot q+b^q_{e_{jk}}))
\end{equation}
Here, $e_{jk}$ indicates the type of dependency relation between node $j$ and its parent node $k$. 

When propagating the attention map encoding $att_j$, we use convolution to project $att_j$, as shown in Figure~\ref{fig:edge}. There are a total of $22$ relation types that reside in the dependency parse tree for questions in the CLEVR dataset; thus, we have $e_{jk} \in [1,22]$.

\subsection{The proposed PTGRN model}
Given the tree-structured layout of the dependency tree, our PTGRN module is sequentially used on each word node to mine visual evidence and integrate features of its child nodes from bottom to top, and then it predicts the final answer at the root of the tree.
Formally, each PTGRN module can be written as
\begin{equation}
\begin{aligned}
{att}_{j}& = f_{a}(\{m^a_{ij}\}_{e_{ij} \in E}, v, w), \\
v_j &= {att}_{j} * v, \\
{h}_{j}& = f_{h}(\{m^h_{ij}\}_{e_{ij} \in E}, v_j), \\
{h}^{att}_{j}& = f^{att}_{h}(\{m^a_{ij}\}_{e_{ij} \in E}, att_j), \\
m^{h}_{jk}& = f^h_{e_{jk}}({h}_{j}, q), \\
m^{a}_{jk}& = f^a_{e_{jk}}({h}^{att}_{j}, q)
\end{aligned}
\label{eq:global}
\end{equation}
We process each node by postorder traversing on the dependency tree. The type of edge indicates whether a node serves as a modifier, which can modify their parent node by referring to a more specific object, or as a subject/object of its predicate parent node. We thus pass both the attention map and the hidden representation of a node to its parent based on the edge such that the parent node can generate a more precise attention map as ${att}_{j}$ or integrate the features of child nodes to enhance the representation given the predicate word. 
After propagating through all word nodes, the output message of the root node $[m^h_{root},m^a_{root}]$ is used to predict the answer. We perform global max pooling on the encoded attention map $m^a_{root}$ and concatenate it with $m^h_{root}$. This concatenated feature is passed through a multilayer perceptron with three layers to predict the final answer $y$.

Our model is stacked by a list of node modules with a tree-structured layout. Weights are shared across all node modules. The entire model can be trained in an end-to-end manner with only the supervision signal $y$.

\section{Experiment}
\begin{figure*}[h]
	\centering
	\includegraphics[width=1.0\textwidth]{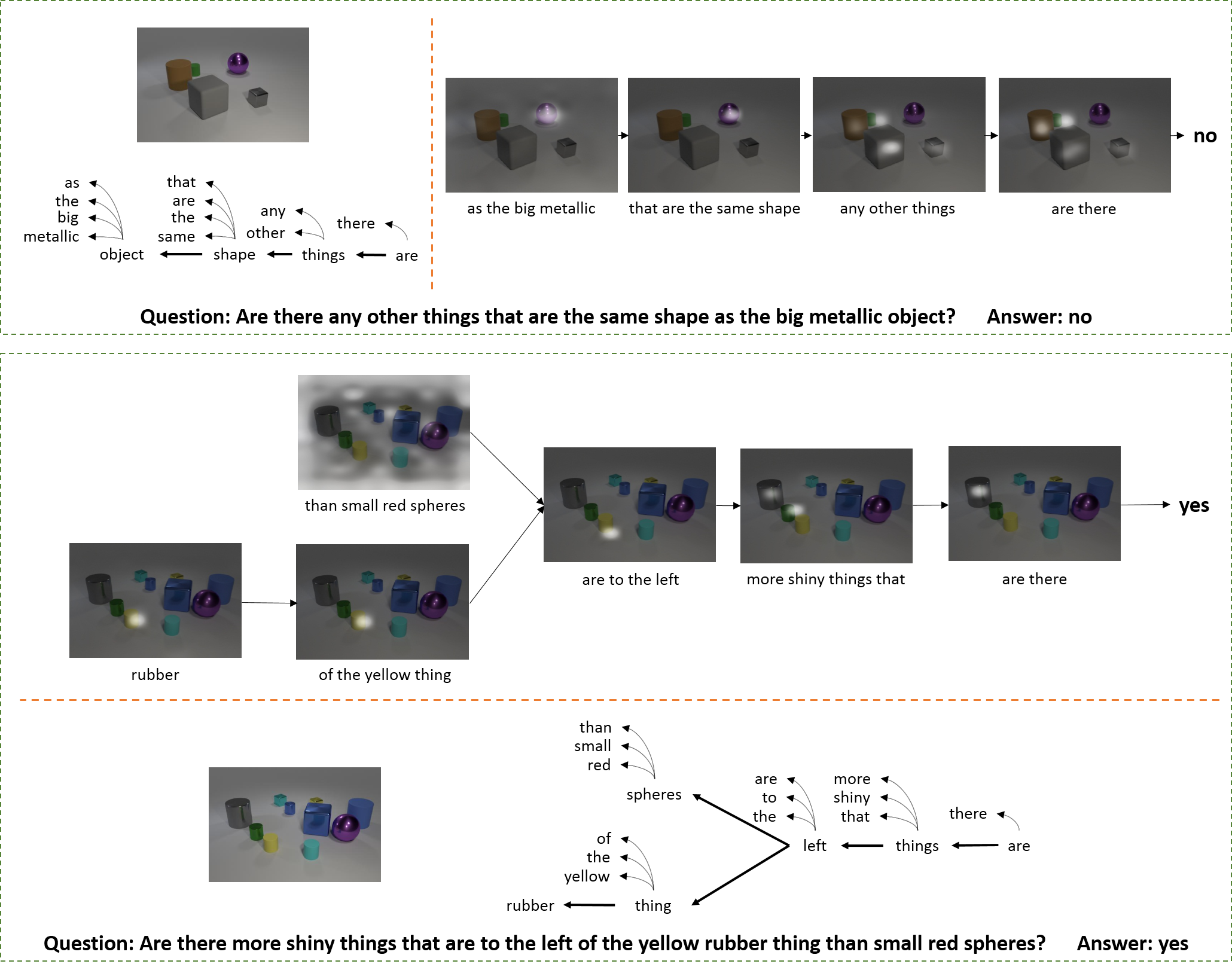}
	\caption{Two examples of the dependency trees of questions and corresponding regions attended by our model at each step on the CLEVR dataset. The questions are shown on the bottom. The input images and dependency parse trees are shown on the left and lower parts. The arrows in the dependency tree are drawn from the head words to the dependent words. The curved arrows point to pruned leaf words that are not nouns. Thus, the word ``are'' is the root node for both examples. 
	}
	\label{fig:clevr}
\end{figure*}
We validate the effectiveness and interpretation capability of our models on both synthetic datasets (i.e., CLEVR and FigureQA) that focus mainly on relation reasoning and natural image datasets (i.e., VQAv2).
\subsection{Datasets}
\textbf{CLEVR}~\cite{clevr} is a synthesized dataset with $100,000$ images and $853,554$ questions. The images are photorealistic rendered images with objects of random shapes, colors, materials and sizes. The questions are generated using sets of functional programs, which consist of functions that can filter certain colors and shapes or compare two objects. Thus, the reasoning routes required to answer each question can be precisely determined by the underlying function program. Unlike natural image datasets, this dataset requires a model capable of reasoning on relations to answer the questions. 

\textbf{FigureQA}~\cite{figureqa} is also a synthesized dataset. This dataset contains $100,000$ images and $1,327,368$ questions for training. In contrast to CLEVR, the images are scientific-style figures. The dataset includes five classes: line plots, dot-line plots, vertical and horizontal bar graphs, and pie charts. The questions also concern various relationships between elements in the figures, such as the maximum, the area under the curve, intersections, etc. Thus, this dataset also requires the VQA model to perform relational reasoning on the plot elements.

The \textbf{VQAv2}~\cite{balanced_vqa_v2} is a widely used VQA benchmark on natural images. It contains $204,721$ natural images from COCO~\cite{Lin2014} and $1,105,904$ free-form questions. Compared with its first version~\cite{VQA}, this dataset focuses on reducing dataset biases through balanced pairs: for each question, there are a pair of images that have different answers.

\begin{table*}[th!]
	\centering
	\caption{Comparison of question answering accuracy on the CLEVR dataset. The performances of question types \emph{Exist}, \emph{Count}, \emph{Compare Integer}, \emph{Query}, and \emph{Compare} are reported in each column. Methods with $*$ are trained with extra program layout annotations. }
	\label{table:clevr}
	% \resizebox{\textwidth}{!}{
	%NS-VQA (90 programs)~\cite{NSVQA} & 87.4 & 64.5 & 53.7 & 77.4 & 79.7 & 74.4\\
	\begin{tabular}{|l|cc|ccc|c|}
		\hline
		Method & Exist & Count & \begin{tabular}[c]{@{}c@{}}Compare\\Integer\end{tabular} & \begin{tabular}[c]{@{}c@{}}Query\\Attribute\end{tabular} & \begin{tabular}[c]{@{}c@{}}Compare\\Attribute\end{tabular} & Overall \\
		\hline
		LBP-SIG~\cite{SAVQA} & 79.6& 61.3 & 80.7 & 88.6 & 76.3 & 78.0 \\
		N2NMN scratch~\cite{e2emn}& 72.7& 55.1& 78.5 & 83.2 & 50.9 & 69.0 \\
		N2NMN cloning expert*~\cite{e2emn} & 83.3 & 63.3 & 80.3 & 87.0 & 78.5 & 78.9 \\
		N2NMN policy search*~\cite{e2emn} & 85.7& 68.5 & 84.9 & 90.0 & 88.7 & 83.7\\
		PE-semi-9K*~\cite{inferring2017} & 89.7& 79.7 & 79.7 & 92.6 & 96.0 & 88.6 \\
		PE-Strong*~\cite{inferring2017} & 97.7 & 92.7 & 98.7 & 98.1 & 98.9 & 96.9\\
		RN~\cite{RelNet} & 97.8 & 90.1 & 93.6 & 97.9 & 97.1 & 95.5\\
		FiLM~\cite{FiLM} & 99.2 & 94.5 & 93.8 & 99.2 & 99.0 & 97.6\\
		MAC~\cite{mac} & 99.5 & 97.1 & 99.1 & 99.5 & 99.5 & 98.9\\
		TbD+reg+hres*~\cite{TbD} & 99.2 & 97.6 & 99.4 & 99.5 & 99.6 & 99.1\\
		NS-VQA*~\cite{NSVQA} & 99.9 & 99.7 & 99.9 & 99.8 & 99.8 & 99.8\\
		ACMN & 94.2 & 81.4 & 81.6 & 90.5 & 97.1 & 89.3 \\
		PTGRN & 97.9 & 91.8 & 95.2 & 95.5 & 98.5 & 95.5 \\
		\hline
	\end{tabular}
	% }
\end{table*}
\subsection{Implementation details}
For the CLEVR dataset, we employ the same settings used in \cite{SAVQA,clevr} to extract image features and word encodings. We first resize all images to $224 \times 224$, and then we extract the conv4 feature from ResNet-101 pretrained on ImageNet. The resulting $1024 \times 14 \times 14$ feature maps are concatenated with a $2$-channel coordinate map, which is further fed into a single $3 \times 3$ convolution layer. The resulting $128 \times 14 \times 14$ feature maps are also concatenated with the $2$-channel coordinate map and then passed through our PTGRN module. We encode the questions using a bidirectional GRU with $512$-d hidden states for both directions. The hidden vector of Bi-GRU at the corresponding word position is extracted to be this word's encoding $w$. The hidden representations of gated residual composition modules are $128$-d for both $h_j$ and $h^{att}_j$. 
The messages generated by parse-tree-guided propagation module modules are also $128$-d for both the attention map and the hidden representation. The three-layer MLP has output sizes of $512$, $1024$ and $29$, where $29$ is the number of candidate answers. 

\begin{figure*}[h]
	\centering
	\includegraphics[width=1.0\textwidth]{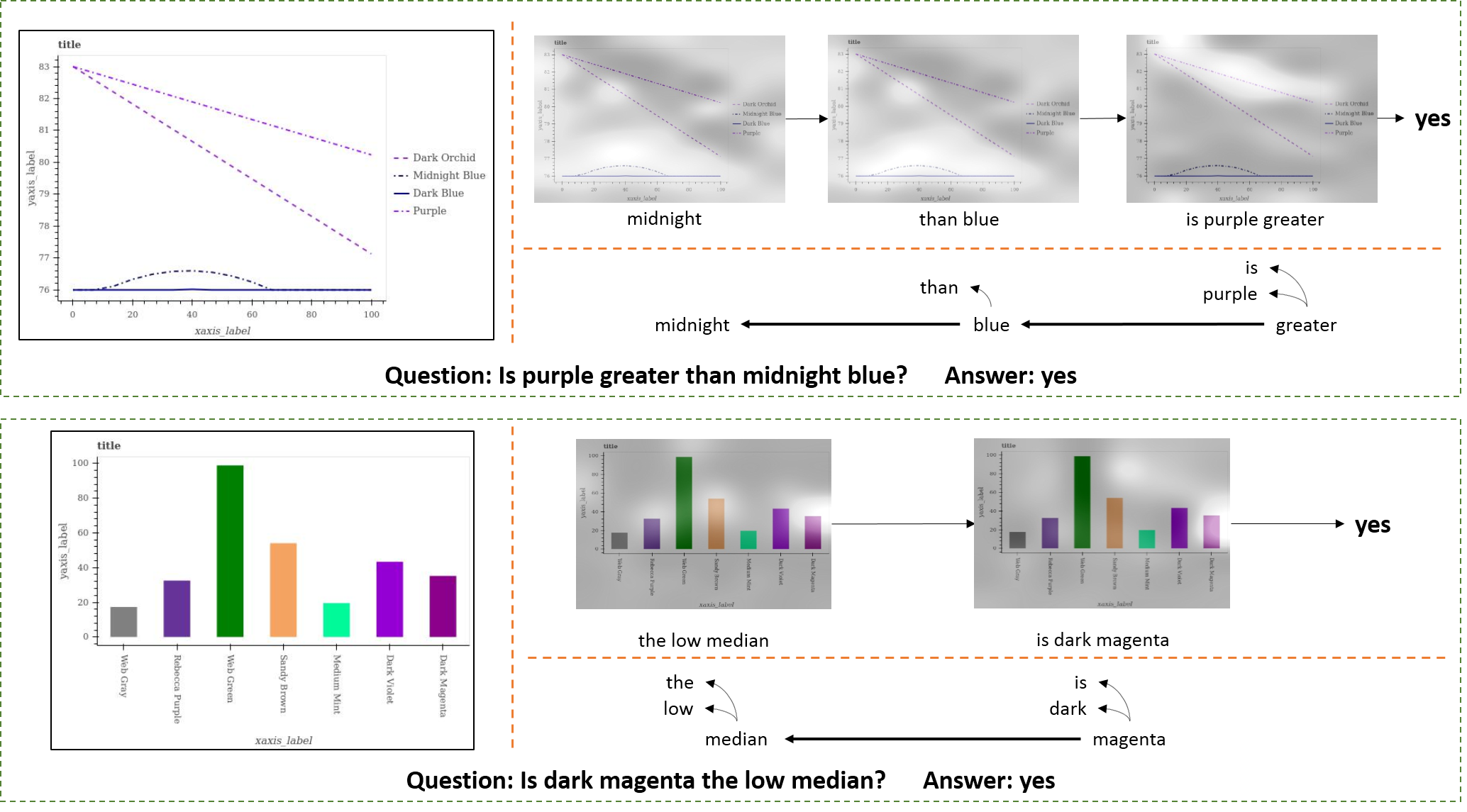}
	\caption{Two examples of the dependency trees of questions and the corresponding regions attended by our model at each step on the FigureQA dataset.}
	\label{fig:figureqa}
\end{figure*}

We resize the images in FigureQA to $256 \times 256$. Then, we use a five-layer CNN to extract the image features. Each layer has a $3 \times 3$ kernel and a stride of $2$. The dimensions of the feature map in the first four layers are $64$, and the last layer has $128$-channel outputs. The resulting feature map is concatenated with the $2$-d coordination map. Thus, the image feature map has a size of $130 \times 8 \times 8$. The hidden output representations of gated residual composition modules and propagation modules are $128$-d. The words in a question are first embedded as a $200$-d vector, and then the whole question is encoded by a bidirectional GRU, which has $1024$-d hidden units in both directions. The word encoding vector is represented by the GRU hidden vector at its corresponding position. 

For the VQAv2 dataset, we use the image features provided by a bottom-up attention network~\cite{Anderson2017up-down}. The bottom-up attention network~\cite{Anderson2017up-down} detects $36$ objects for each image and extracts $2048$-d regional features for each object. The word vectors are extracted from a $512$-d bidirectional LSTM at corresponding positions. The hidden representation and attention map encoding are $1024$-d and $128$-d, respectively. 
Since attention is performed on $36$ objects instead of the feature map, we use a fully connected layer rather than a convolutional layer to encode the attention results. Specifically, we concatenate the $36$-d attention weights with corresponding $4$-d bounding box coordinates, then flatten it to a $36*5$-d vector, and finally fed the $180$-d vector into a fully-connected layer to produce the $128$-d attention map encoding.

For CLEVR and FigureQA dataset, the model is trained with the Adam optimizer~\cite{adam}. The base learning rate is $0.0003$ for CLEVR, and it is $0.0001$ for FigureQA. The batch size is $64$. The weight decay, ${\beta}_1$ and ${\beta}_2$ are $0.00001$, $0.9$, and $0.999$, respectively. We train our model on the training split, and then we evaluate on the validation or test split.
For VQAv2 dataset, we train the model on VQAv2 on training and validation split with Adamax~\cite{adam} with a learning rate of $0.001$ and a batch size of $128$.
\begin{table}[t] \centering
	\center
	\caption{Comparison of question answering accuracy on FigureQA validation and test sets that have alternative color schemes.}
	\label{table:figureqa}
	\begin{tabular}{|l|*{2}{c|}}
		\hline
		Model & Val & Test \\
		\hline
		Text only & 50.01 & 50.01\\
		CNN+LSTM & 56.16 & 56.00\\
		CNN+LSTM on VGG-16 features & 52.31 & 52.47\\
		RN & 72.54 & 72.40\\
		Ours & 86.25 & 86.23 \\ %88.23
		\hline
	\end{tabular}
\end{table}

\subsection{Comparison with state-of-the-art models}
\subsubsection{CLEVR dataset}
Table~\ref{table:clevr} shows the performances of different works on the CLEVR test set. The previous end-to-end modular network~\cite{e2emn} and program execution engine~\cite{inferring2017} are referred to as N2NMN and PE, respectively. Both approaches use functional programs as the ground-truth layout, and they train their question parser in a sequence-to-sequence manner with strong supervision. They also have variants that are trained using signals with semisupervision or no supervision. ``N2NMN scratch'' indicates the end-to-end modular network without layout supervision and ``N2NMN cloning expert'' shows the results of models trained with full supervision. ``N2NMN policy search'' provides this model's best results if it further trains the parser from ``N2NMN cloning expert'' with RL. 
As shown, our model outperforms the previous models by a large margin without using a dataset-specific layout, thus showing the good generalizability of PTGRN. PTGRN also surpasses the program execution engine~\cite{inferring2017} variant trained with semisupervision (as ``PE-semi-9K''). 
% MAC~\cite{mac} NS-VQA~\cite{NSVQA} TbD+reg+hres~\cite{TbD}
Furthermore, PE-Strong~\cite{inferring2017}, NS-VQA~\cite{NSVQA} and TbD+reg+hres~\cite{TbD} used extra program layouts as additional supervision signals. RN~\cite{RelNet} and FiLM~\cite{FiLM} are black-box models that lack interpretability. PTGRN not only obtains comparable accuracy, but also provides more explicit reasoning results without dataset-specific layout supervision.

``ACMN'' shows the results of our preliminary work. It performs structured reasoning along the dependency tree but has three differences. 1) Rather than learning a forget gate, it drops the attention map or hidden representation according to whether the edge is a clausal predicate relation or a modifier relation. 2) It performs fusion between hidden and question encodings independent of edge type, but each node shares only weights with other nodes at the same height. 3) It propagates the attention map without extra encoding, and the attention module masks the image feature adversary based on the children's attention maps. As shown, our proposed modules significantly improve the performance by $6.2\%$. 

Figure~\ref{fig:clevr} shows the promising intermediate reasoning results achieved by PTGRN. The images and the dependency parse trees are shown on the left and bottom. The attention map that our model obtained at each tree node is displayed on the right and top. The first example shows that our model can first locate the ``big metallic object'', while the phrase ``same shape'' attends the same region. Later, our model attends all objects except the metallic object given the phrase ``any other things'', and the phrase ``are there'' extracts the attended objects' features and predicts the answer ``no''. The second example first locates the ``yellow rubber'' thing, and it attends nothing given the phrase ``small red spheres'' because there is no such object in the image. Then, it sequentially attends the ``left'' and ``shiny things'' based on the yellow rubber object and predicts the answer ``yes''.

\subsubsection{FigureQA dataset}
Table~\ref{table:figureqa} presents the comparisons of our model with prior works on the FigureQA dataset. The baseline methods ``Text only'', ``CNN+LSTM'', ``CNN+LSTM on VGG-16 features'' and ``RN~\cite{RelNet}'' were originally reported in~\cite{figureqa}. As shown, PTGRN outperforms all baseline methods by a large margin, including the relational network, which has achieved great performance on the CLEVR dataset. This result demonstrates the generalizability of our model across different datasets.
Figure~\ref{fig:figureqa} shows the reasoning route on the FigureQA dataset. The first question queries whether the plot of purple is greater than that of midnight blue. Our model successfully locates midnight blue in the first two steps, and then it locates the dotted purple line and predicts the answer. The second example first locates several bars that are ``median'', and then it attends the ``dark magenta'' bar to predict that it is the low median. 
\begin{table}[!tp] \centering %\small%\footnotesize %
	\center
	\caption{Question answering accuracy on VQAv2 test-dev and test-std.}
	\label{table:vqav2}
	\resizebox{\columnwidth}{!}{%
		\begin{tabular}{|c|*{4}{c}|*{4}{c}|}
			\hline
			& \multicolumn{4}{c|}{test-dev} & \multicolumn{4}{c|}{test-std} \\
			Method & All & Yes/no & Numb. & Other & All & Yes/no & Numb. & Other\\
			\hline
			Bottom-Up~\cite{vqav2winner} & 65.32 & 81.82 & 44.21 & 56.05 & 65.67 & 82.20 & 43.90 & 56.26 \\
			Counter~\cite{learn2count} & 68.09 & 83.14 & 51.62 & 58.97 & 68.41 & 83.56 & 51.39 & 59.11 \\
			BAN+Glove+Counter~\cite{BAN} & 70.04 & 85.42 & 54.04 & 60.52 & 70.35 & - & - & - \\
			ACMN & 63.81 & 81.59 & 44.18 & 53.07 & 64.05 & 81.83 & 43.80 & 53.22 \\
			PTGRN & 67.07 & 83.69 & 48.47 & 57.09 & 67.26 & 83.67 & 48.22 & 57.33 \\
			\hline
		\end{tabular}
	}
	\vspace{-4mm}
\end{table}

\begin{figure*}[hbtp]
	\centering
	\includegraphics[width=0.45\textwidth]{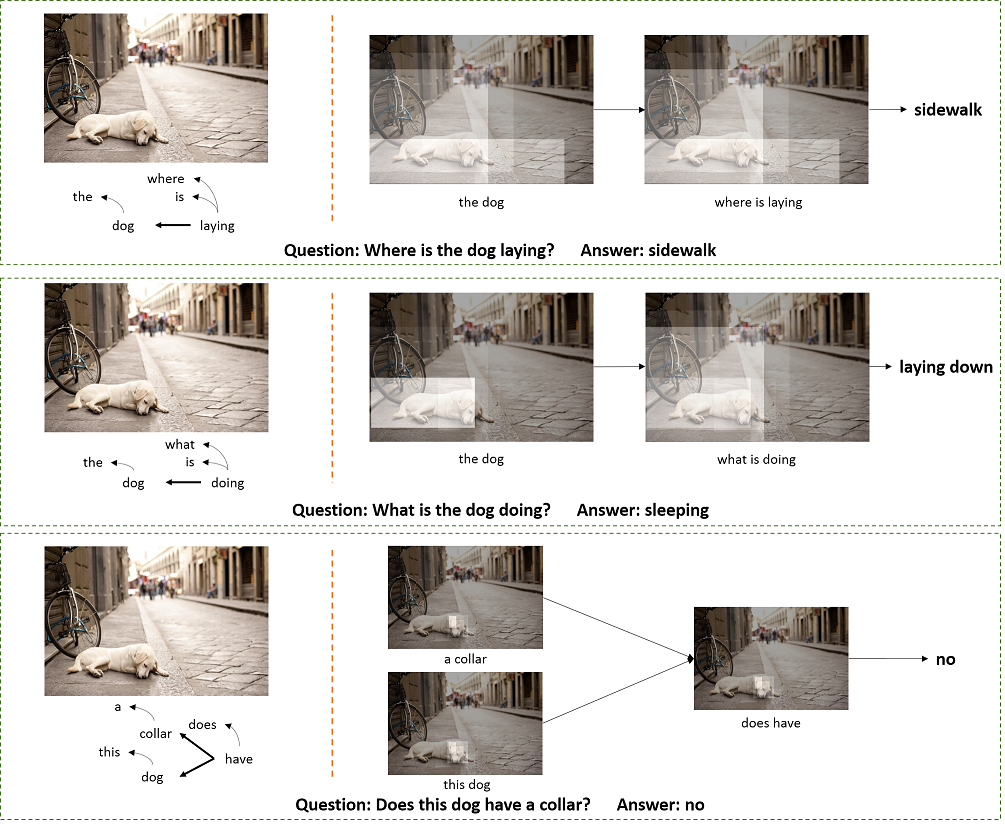}
	\includegraphics[width=0.45\textwidth]{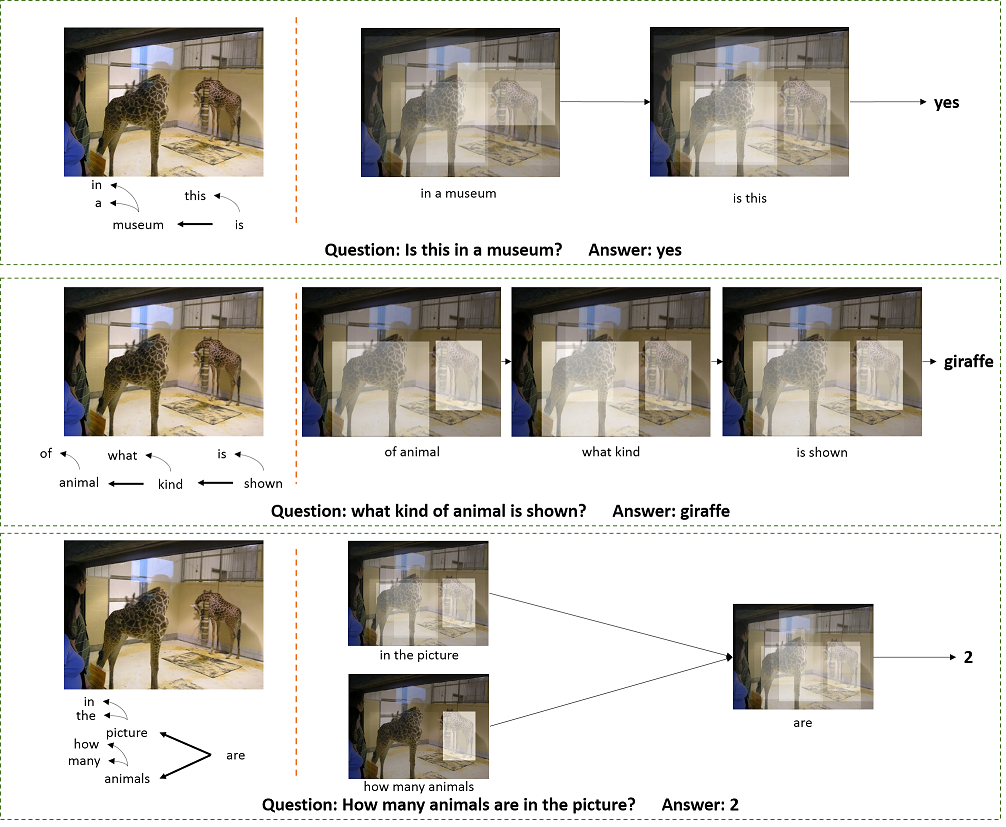}
	\caption{Examples of the dependency trees of questions and the corresponding regions attended by our model at each step on the VQAv2 dataset.}
	\label{fig:vqav2}
\end{figure*} 
\subsubsection{VQAv2 dataset}
We compare our model with state-of-art methods on the VQAv2 dataset, The results are shown in Table~\ref{table:vqav2}, and the intermediate reasoning results are shown in Figure~\ref{fig:vqav2}.

Compared with bottom-up~\cite{vqav2winner} baseline, our model gains improvement on overall accuracy and gains large improvement on ``Number'' questions. 
Our model don't outperform other state-of-the-art model for the following two reasons.
First, current state-of-the-art models use many techniques such as multiple glimpse, stacked attention and data augmentation to achieve high accuracy. These techniques can be applied generally and thus cannot determine the contribution of a certain model. 
Second, our model performs reasoning on dependency parse trees and handles the compositional reasoning of words in the questions, but it gains no advantage on reasoning with a single word. 
However, as shown in Figure~\ref{fig:vqav2}, most questions in VQAv2 are structurally simple. They do not require compositional reasoning of elements in the image but require complex commonsense reasoning that is outside the image. One of the most common examples is the word ``why''. To give an interpretable reasoning process, it may need to perform multistep inference on a commonsense knowledge graph, which is not included in the VQAv2 dataset.
%Learning the knowledge graph purely from data is a black-box process that is currently being researched.
%This work intends to address the problem of interpretable and compositional reasoning and to use a similar attention mechanism as other methods for single word reasoning, leading to similar performance on VQAv2.

\begin{figure*}[h]
	\centering
	\includegraphics[width=1.0\textwidth]{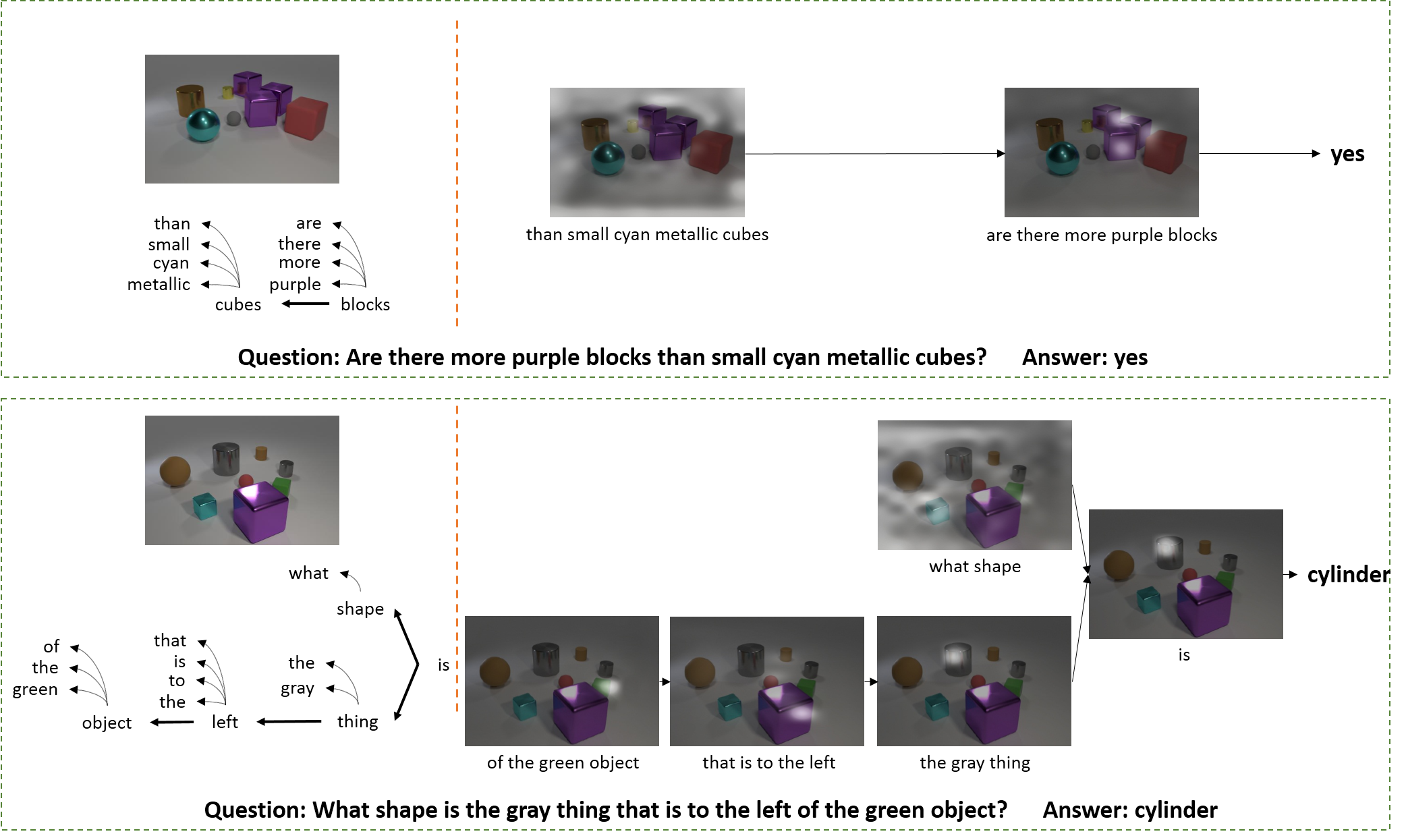}
	\caption{Two examples of the dependency trees of questions and corresponding regions attended by our model at each step on the CLEVR-CoGen test set. }
	\label{fig:clevr_cogen}
\end{figure*}

\subsection{CLEVR composition generalization test}
\begin{table}[t] \centering
	\caption{Comparisons of question answering accuracy on the CLEVR-CoGenT validation set. Each method is trained on condition A only and evaluated on both condition A and condition B.}
	\label{table:clevr_CoGenT}
	\begin{tabular}{|r|c|c|}
		\hline
		& \multicolumn{2}{c|}{Train A} \\
		Model & \multicolumn{1}{c}{A} & \multicolumn{1}{c|}{B} \\
		\hline
		CNN+LSTM+SA & 80.3 & 68.7 \\
		PG+EE (18K prog.) & 96.6 & 73.7 \\
		CNN+GRU+FiLM & 98.3 & 75.6 \\
		CNN+GRU+FiLM 0-Shot & 98.3 & 78.8 \\
		\hline
		Ours & 97.35 & 83.50 \\
		\hline
	\end{tabular}
\end{table}
The CLEVR composition generalization test (CLEVR-CoGenT)~\cite{clevr} was proposed to investigate the composition generalizability of a VQA model. This dataset contains synthesized images and questions similar to CLEVR, but it has two conditions: in condition A, all cubes are gray, blue, brown, or yellow, and all cylinders are red, green, purple, or cyan; in condition B, cubes and cylinders swap color palettes. Thus, one model cannot achieve good performance on condition B by simply memorizing and overfitting the samples in condition A.

We report the accuracy of our model in Table~\ref{table:clevr_CoGenT}. ``Ours'' represents the model that we trained on the CLEVR dataset with the same settings and hyperparameters. We train it on a training set that meets condition A and evaluate it on a validation set that meets condition A and condition B. Our model achieves $97.35\%$ accuracy on condition A, and it achieves $83.50\%$ accuracy on condition B without being fine-tuned on the alternative color scheme set B. Our method achieves higher accuracy on condition B, while the accuracy on condition A is similar to PE~\cite{inferring2017} and FiLM~\cite{FiLM}. This result demonstrates that our model has better composition generalizability.

In Figure~\ref{fig:clevr_cogen}, we show our model's reasoning routes on the CLEVR-CoGenT condition B validation set. We display the image and parse tree on the left, and the attention map at each step is shown on the right. The first questions query the number of purple blocks and cyan metallic cubes. When there are no ``cyan metallic cubes'', the first step attends nothing, while the second step successfully attends the three purple blocks. Note that there is no purple block in the training set, and our model correctly distinguishes the purple blocks from the red block. The second example follows the parsed tree and successively attends the ``green object'', its ``left'' and the ``gray thing''. Gray cylinders are not present in the training set, but our model can locate the color ``gray'' and then predict the answer. These examples show that our model can locate the color and the object separately and that it possesses composition generalizability to a certain extent.
\begin{figure*}[hbtp]
	\centering
	\includegraphics[width=0.95\textwidth]{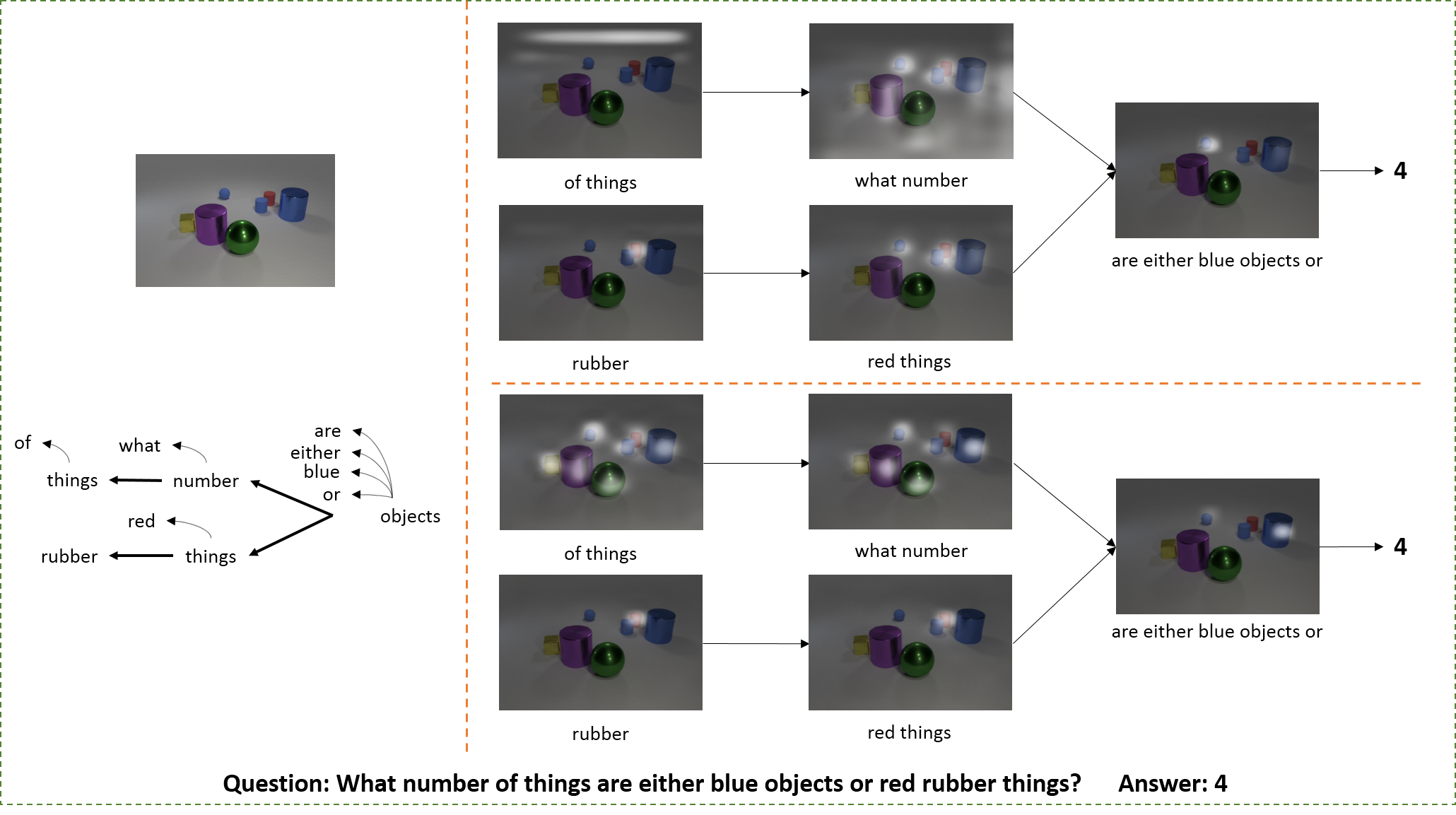}
	\includegraphics[width=0.95\textwidth]{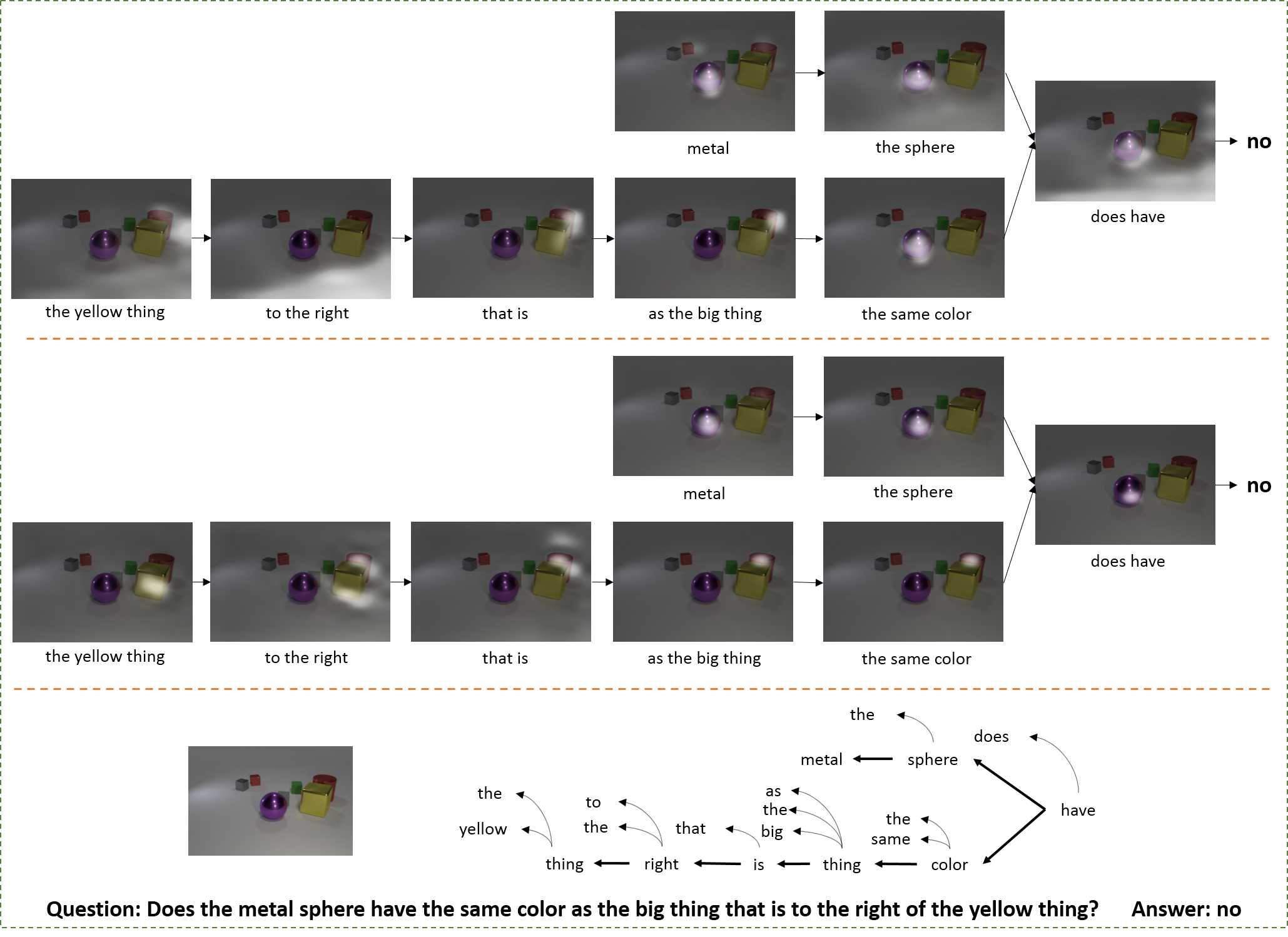}
	\caption{Quality comparison of the model with and without GRU and the propagation module}
	\label{fig:clevr_cmp}
\end{figure*} 
\subsection{Ablation studies} 
\begin{table*}[th!]
	\centering
	\caption{CLEVR validation accuracy for ablations. We incrementally replace the components of the baseline model with redeveloped modules and report their results on each row.}
	\label{table:clevr_as}
	\resizebox{\textwidth}{!}{
		\begin{tabular}{|l|cc|ccc|cccc|cccc|c|}
			\hline
			& & &\multicolumn{3}{c|}{Compare Integer} & \multicolumn{4}{c|}{Query} & \multicolumn{4}{c|}{Compare} & \\
			\hline
			Method &Exist &Count & Equal& Less& More & Size& Color&Material&Shape&Size&Color& Material& Shape & Overall\\ 
			\hline
			ACMN & 94.20 & 80.46 & 74.21 & 88.98 & 83.62 & 93.10 & 86.42 & 92.16 & 89.72 & 97.17 & 96.54 & 95.94 & 96.64 & 89.01 \\
			baseline & 91.38 & 75.62 & 76.24 & 89.00 & 85.26 & 92.71 & 86.20 & 92.33 & 89.18 & 98.29 & 96.06 & 97.43 & 97.05 & 87.62 \\
			\hline
			+Edge & 93.98 & 80.18 & 72.18 & 87.55 & 84.86 & 93.37 & 86.61 & 92.35 & 90.36 & 98.92 & 98.01 & 98.20 & 97.72 & 89.28 \\
			+Edge+GRU & 95.23 & 82.62 & 83.93 & 92.04 & 91.89 & 92.61 & 85.92 & 92.23 &89.44 &98.56 &98.37 &97.88 &97.69 & 90.44 \\
			+Edge+ConcatAtt & 96.38 & 85.51 & 86.28 & 95.37 & 93.81 & 95.99 & 95.22 & 95.37 & 94.48 & 98.74 & 97.59 & 98.12 & 97.66 & 93.36 \\
			+Edge+GRU+ConcatAtt & 95.81 & 86.31 & 85.44 & 94.90 & 93.13 & 96.03 & 96.12 & 95.58 & 94.06 & 98.72 & 98.32 & 98.23 & 97.88 & 93.53 \\
			\hline
			+Edge+GRU+GRUAtt & 96.80 & 87.50 & 89.00 & 96.04 & 95.33 & 96.76 & 96.18 & 96.31 & 95.11 & 99.39 & 98.76 & 98.27 & 98.03 & 94.42 \\
			+Edge+GRU+ConvGRUAtt & 97.83 & 91.78 & 92.08 & 96.00 & 96.46 & 95.90 & 94.00 & 95.46 & 96.15 & 99.30 & 98.32 & 98.52 & 98.19 & 95.42 \\
			\hline
		\end{tabular}
	}
\end{table*}
\begin{table*}[th!]
	\centering
	\caption{CLEVR validation accuracy of PTGRN with different word encoding and provided layout.}
	\label{table:clevr_as2}
	\resizebox{\textwidth}{!}{
		\begin{tabular}{|l|cc|ccc|cccc|cccc|c|}
			\hline
			& & &\multicolumn{3}{c|}{Compare Integer} & \multicolumn{4}{c|}{Query} & \multicolumn{4}{c|}{Compare} & \\
			\hline
			Method &Exist &Count & Equal& Less& More & Size& Color&Material&Shape&Size&Color& Material& Shape & Overall\\
			\hline
			PTGRN (lookup) & 92.66 & 78.03 & 71.19 & 86.14 & 83.68 & 93.22 & 86.62 & 92.92 & 90.24 & 94.78 & 95.52 & 92.64 & 93.11 & 87.75 \\
			PTGRN (LSTM) & 95.60 & 86.93 & 87.55 & 93.35 & 93.49 & 94.61 & 88.06 & 93.91 & 91.92 & 97.00 & 95.64 & 91.19 & 95.27 & 91.82 \\
			PTGRN (GT layout) & 98.67 & 97.70 & 97.53 & 89.79 & 89.43 & 98.63 & 98.04 & 97.81 & 97.62 & 99.91 & 98.92 & 98.78 & 98.47 & 96.69 \\
			PTGRN & 97.83 & 91.78 & 92.08 & 96.00 & 96.46 & 95.90 & 94.00 & 95.46 & 96.15 & 99.30 & 98.32 & 98.52 & 98.19 & 95.42 \\
			\hline
		\end{tabular}
	}
\end{table*}

We first compare the performance with and without the redeveloped components in our model. We incrementally add the modules to the baseline model, and we show the accuracy on the CLEVR validation set in Table~\ref{table:clevr_as}.

\textbf{ACMN} \quad
The adversarial composition modular network (ACMN) presented in the preliminary work performs structured reasoning along a general dependency parse tree. In each node, it first mines local visual evidence with the adversarial attention module: the image feature is masked by $ReLU(1-\tilde{att})$, where $\tilde{att}$ is the sum of the children's attention map; then, the masked image feature is fused with word embeddings and convolved to a $1$-d attention map $att$. Then, the attended local image feature $v$ is composed with the sum of the children's hidden representation $\tilde{h}$: it generates the residual by applying a fully connected layer on concatenated $[v,\tilde{h}]$; then, the residual is added to $\tilde{h}$ and fused with the question encoding, and the current hidden $h$ is obtained as the result. Finally, the attention map $att$ is propagated to its parent if the edge type belongs to a modifier relation, or hidden $h$ is propagated if the edge is a clausal predicate relation. Each node will share weights with other nodes at the same height.

\textbf{Baseline model} \quad
Compared with ACMN, the baseline model has the following changes: 1) each node now shares its weights with other nodes across different heights in the parsed tree and 2) the adversarial attention module is replaced by the proposed one described in section~\ref{attention_module}. To perform ablation studies on PTGRN, we replace each module in the baseline model with our proposed components and evaluate them incrementally.

\textbf{Edge-dependent propagation} \quad
``+Edge'' performs extra edge-dependent feature transformations between children and parents. These changes increase the performance by $0.27\%$ compared to that of ACMN and make the model easier to apply to different datasets without adjusting the maximum tree height. Moreover, it is more reasonable to reuse the same module to perform the same tasks, and the accuracy is increased by $1.66\%$ over the baseline model.

\textbf{Gated residual} \quad
We investigate the effect of the gated residual module by comparing the accuracy with that of the ``+Edge'' model. 
The ``+Edge'' model constructs the mined visual evidence $v_j$ and its children nodes' message $\tilde{h}$ by concatenating them and performing a linear transform with a fully connected layer, while the ``+Edge+GRU'' model replaces it with a GRU, where $v_j$ corresponds to the current input, and the message is the memory. Adding a GRU improves the accuracy by $1.16\%$. This result suggests that the GRU module stabilized the recurrent process and improved the performance. 
If the attention map is encoded into a hidden representation, then whether the incoming message is gated has little effect on performance. With the forgotten gate, the accuracy increases from $93.36\%$ to $93.53\%$, as shown in ``+Edge+ConcatAtt'' and ``+Edge+GRU+ConcatAtt''.

\textbf{Attention map encoding} \quad
There are multiple methods for encoding the attention map as a hidden representation. We evaluate several methods and report their answering accuracy in Table~\ref{table:clevr_as}.

The first method is to flatten the attention map, apply a linear layer, and concatenate the resulting vector to hidden and propagate it to the parent. In the case of the CLEVR dataset, the $14*14$ attention map $att_j$ is flattened to a $196$-d vector and projected to $128$-d feature $h^{att}_j$ with two fully connected layers. Then, it is concatenated with the mined visual evidence $v_j$. The accuracies are reported as ``+Edge+ConcatAtt'' and ``+Edge+GRU+ConcatAtt'', where the first one composes the current hidden $[v_j, h^{att}_j]$ and input hidden $\tilde{h}$ by concatenating them, as in the preliminary work. In ``+Edge+GRU+ConcatAtt'', we compose the $[v_j, h^{att}_j]$ and $\tilde{h}$ with a GRU-unit, as described in section~\ref{GRCM}. Both methods pass the original attention map $att_j$ to its parent. ``+Edge+ConcatAtt'' achieves $93.36\%$ overall accuracy. ``+Edge+GRU+ConcatAtt'' achieves $93.53\%$, which is slightly higher than the $93.36\%$ obtained by the model without GRU. From the accuracy per question type, we observe that ``+Edge+ConcatAtt'' performs slightly better on ``compare integer'' questions. These results indicate that updating the encoded attention map and the hidden representation together will harm the counting ability of our model.

The second method investigates the influence of the separated modules for the attention map. We perform the experiment denoted as ``+Edge+GRU+GRUAtt''. This method also first flattens the attention map to a $196$-d vector. Then, this vector is used to update the input attention feature $\tilde{att}_j$ with another GRU rather than concatenating it with mined visual evidence $v_j$. The output hidden representation of the GRU $h^{att}_j$ will be propagated to its parent node $k$ based on the edge $e_{jk}$. Finally, the encoded attention map to the root node $m^{a}_{root}$ will be concatenated with the hidden $m^{h}_{root}$ and fed into the classifier to predict the answer. It achieves $94.42\%$ overall accuracy on the validation set, which improves the accuracy of the `+Edge+GRU+ConcatAtt'' baseline by $0.89\%$, illustrating that it is necessary to process the attention map and the hidden representation with different weights. This result is also consistent with our preliminary model, which processes the attention map and the hidden representation differently based on whether the edge is a modifier relation or a clausal predicate relation. 

The last approach is our full model, which is reported as ``+Edge+GRU+ConvGRUAtt''. Compared with the second one, this approach utilizes the convolution GRU to preserve spatial information and propagate feature maps across the tree. We perform global max pooling on the feature map $m^{a}_{root}$ and concatenate it with the hidden representation to predict the final answer. It gains $1\%$ over the second approach on the validation set. It is shown that preserving spatial information will lead to better performance.

\textbf{Bidirectional encoding for pruned words} \quad
In our implementation, we pruned leaf nodes that are not nouns to reduce the computational burden. In this way, we can ignore determiner and preposition words such as ``the'' and ``of''. However, this method also prunes some modifier words such as ``big'' and ``metallic'' in Figure~\ref{fig:clevr}. Since the modifiers are near the object words, we use bidirectional LSTM to extract the hidden representations that encode both the modifiers and the object words. We train two variants that use a lookup table and LSTM to extract word encodings. The results are reported in Table~\ref{table:clevr_as2} and denoted as ``PTGRN (lookup)'' and ``PTGRN (LSTM)'', respectively. The lookup table encoding, which did not consider contexts, led to a significant accuracy drop from $95.42\%$ to $87.75\%$. LSTM can encode the modifier words that precede the object, and thus, its performance slightly drops by $3.6\%$.

\textbf{Ground-truth program layout} \quad %Editor: Please use a consistent capitalization and punctuation format for section headings throughout the manuscript. Some journals request a specific style, so please review the journal's guidelines.
We also investigate the effectiveness of our neural modules by providing ground-truth program layouts. The program layouts are used to generate the questions in CLEVR and are built by composing a set of basic functions into a tree/chain structure. Each function performs one type of operation based on object inputs and value inputs. For example, functions with type ``filter\_color'' and value ``blue'' output the subset of blue input objects. 
Given the program layouts, we perform structured reasoning by considering each function as a node and applying our neural module. Given a function $i$, we encode the value input with a lookup table and use the result as a word encoding $w_i$; we use the function type as $e_{ij}$, where $j$ is the subsequent function. The results are denoted as ``PTGRN (GT layout)'' in Table~\ref{table:clevr_as2}. It is shown that our model can achieve results that are good when compared with other methods that used this layout annotation.

\textbf{Quality comparison} \quad
We further compare the reasoning results generated by ACMN and PTGRN in Figure~\ref{fig:clevr_cmp}. The upper part shows the heat map given by ACMN, and the lower part displays the results of PTGRN.
The first question queries the number of specific objects. ACMN correctly attends the ``red rubber'' things, but it locates only one of the blue objects; PTGRN has also found the ``red rubber'' and locates all of the objects given the phrases ``of things'' and ``what number''. Finally, PTGRN attends the two blue objects that are far from the ``red rubber'' at the last step.
The second example illustrates the process of locating the ``big thing'' and the ``metal sphere''. Although both models mined the corresponding objects at each step, PTGRN can pinpoint the location of these objects.

\begin{table}[t]
	\centering
	\caption{The accuracy of attention maps on CLVER validation set.}
	\label{table:attentino}
	\resizebox{\columnwidth}{!}{
		\begin{tabular}{|*5{c|}}
			\hline
			Method & MAC~\cite{mac} & PTGRN & TbD+reg+hres~\cite{TbD} & PTGRN (GT layout)  \\
			\hline
			Accuracy & 32.72 & 64.71 & 77.11  & 91.85  \\			
			\hline
		\end{tabular}
	}
\end{table}

\subsection{Evaluation of interpretability}
We evaluate and compare the interpretability of our model with that of the previous method by calculating how well the generated attention map matches the ground-truth bounding boxes. 
Specifically, the CLEVR dataset provides functional program layouts for each question. The program layouts are composed of functions and some functions output object sets. Given a function output object set and an attention map, we sum the normalized attention weight inside the objects' bounding boxes as the attention accuracy on this function. We evaluate the interpretability of a model by averaging the attention accuracy on all functions in the program layouts.
%For TBD~\cite{TbD}, the layouts are identical to program layouts; thus, we compare the function output with the attention maps generated at corresponding nodes.
For MAC~\cite{mac} and our methods, the reasoning processes are different from the program layouts. Thus, for each function, we use the attention map that achieves highest attention accuracy as the accuracy for this function.
The results are shown in Table~\ref{table:attentino}. Given the ground-truth layouts, our model achieves better results than TbD~\cite{TbD}; Our model also outperforms MAC~\cite{mac} by $31.99\%$ if the ground-truth layouts are not given, demonstrating the interpretability of our model.

\subsection{Perturbed tree structure} 
Since our model relies on the parsed results of the parser, we perform experiments to investigate the impact of incorrectly parsed trees. We randomly distort the generated dependency trees to observe the effect on the final prediction. Specifically, we randomly perturb each dependency relation with a certain probability by replacing its parent node and relation edge with a word that is randomly chosen from the sentence and the possible relations, respectively. The results are listed in Table~\ref{table:perturb}. As shown, the performance dramatically decreases as the number of perturbed relations increases. This result indicates that the parsing results are crucial to the performance.
\begin{table}[t] \centering \footnotesize
	\caption{Accuracy on CLEVR with randomly perturbed dependency trees.}
	\label{table:perturb}
	\begin{tabular}{|*6{c|}}
		\hline
		Percentage of perturb & 0 & 10\% & 30\% & 50\% & 70\% \\
		\hline
		Accuracy & 95.42 & 86.77 & 69.88 & 54.21 & 40.11 \\ %nndep: 95.63
		\hline
	\end{tabular}
\end{table}
\section{Conclusion}
In this paper, we propose a novel parse-tree-guided reasoning network (PTGRN) equipped with an attention module, a gated residual composition module, and a parse-tree-guided propagation module. In contrast to previous works that rely on annotations or handcrafted rules to perform explicit compositional reasoning, our PTGRN model can automatically perform an interpretable reasoning process over a general dependency parse tree based on the question, which can largely broaden its application fields. The attention module encourages the model to attend the local visual evidence, the gated residual composition module can learn to compose and update the knowledge from its child nodes, and the parse-tree-guided propagation module generates and propagates the edge-dependent information from children to their parents. 
Experiments show that PTGRN can achieve state-of-the-art VQA performance and good interpretability without using any specified ground-truth layouts or complicated handcrafted rules.
%Experiments show that PTGRN outperforms previous neural network models without using any specified ground-truth layouts or complicated handcrafted rules.
\ifCLASSOPTIONcaptionsoff
\newpage
\fi
\bibliographystyle{IEEEtran}
\bibliography{egbib}
\begin{IEEEbiography}[{\includegraphics[width=1in,height=1.25in,clip,keepaspectratio]{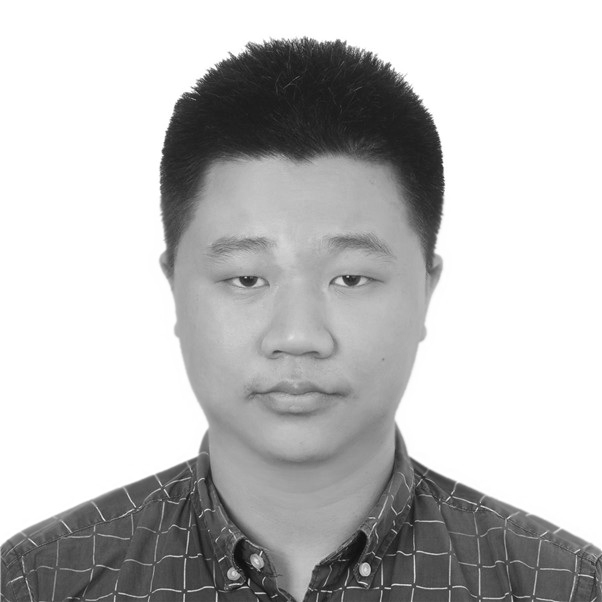}}]{Qingxing Cao}
	Qingxing Cao is currently a postdoctoral researcher in the School of Intelligent Systems Engineering at Sun Yat-sen University, working with Prof. Xiaodan Liang. He received his Ph.D. degree from Sun Yat-Sen University in 2019, advised by Prof. Liang Lin. His current research interests include computer vision and visual question answering.
\end{IEEEbiography}
\begin{IEEEbiography}[{\includegraphics[width=1in,height=1.25in,clip,keepaspectratio]{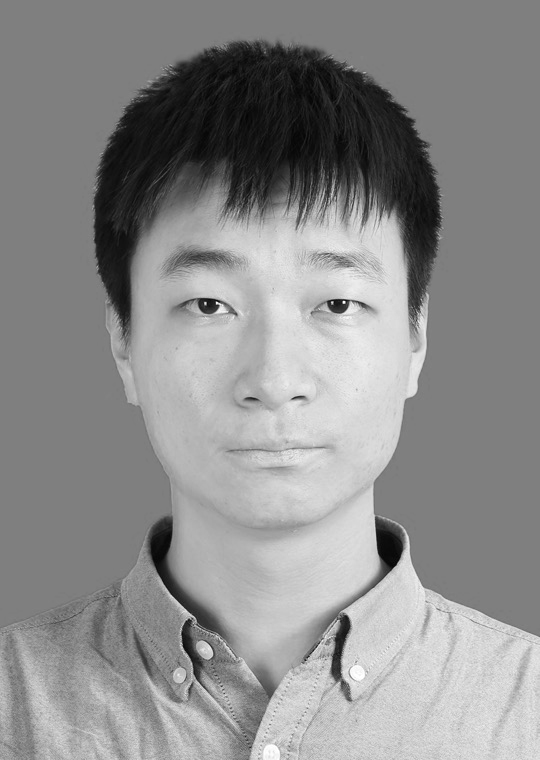}}]{Bailin Li}
	 Bailin Li received his B.E. degree from Jilin University, Changchun, China, in 2016, and the M.S. degree at Sun Yat-Sen University, Guangzhou, China, advised by Professor Liang Lin. He currently leads the model optimization team at DMAI. His current research interests include visual reasoning and deep learning (e.g., network pruning, neural architecture search).
\end{IEEEbiography}
\begin{IEEEbiography}[{\includegraphics[width=1in,height=1.25in,clip,keepaspectratio]{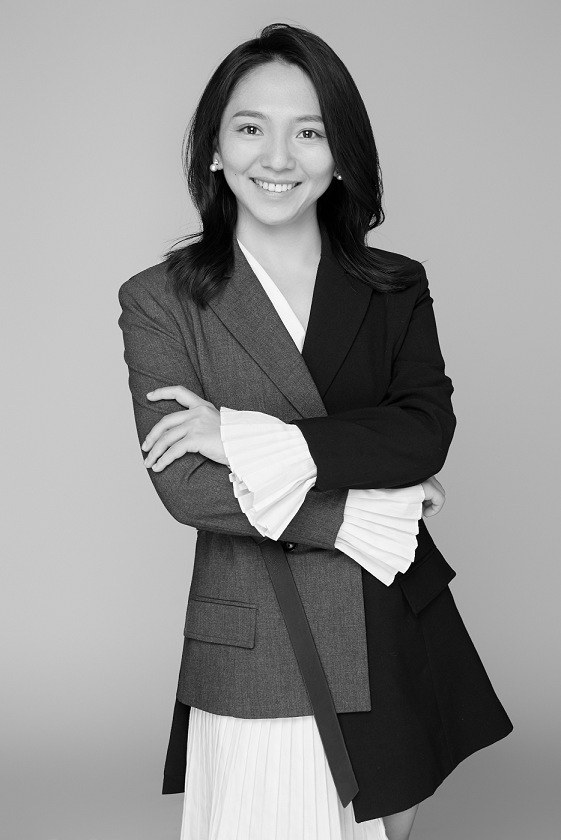}}]{Xiaodan Liang}
	Xiaodan Liang is currently an Associate Professor at Sun Yat-sen University. She was a postdoc researcher in the machine learning department at Carnegie Mellon University, working with Prof. Eric Xing, from 2016 to 2018. She received her PhD degree from Sun Yat-sen University in 2016, advised by Liang Lin. She has published several cutting-edge projects on human-related analysis, including human parsing, pedestrian detection and instance segmentation, 2D/3D human pose estimation and activity recognition.
\end{IEEEbiography}
\begin{IEEEbiography}[{\includegraphics[width=1in,height=1.25in,clip,keepaspectratio]{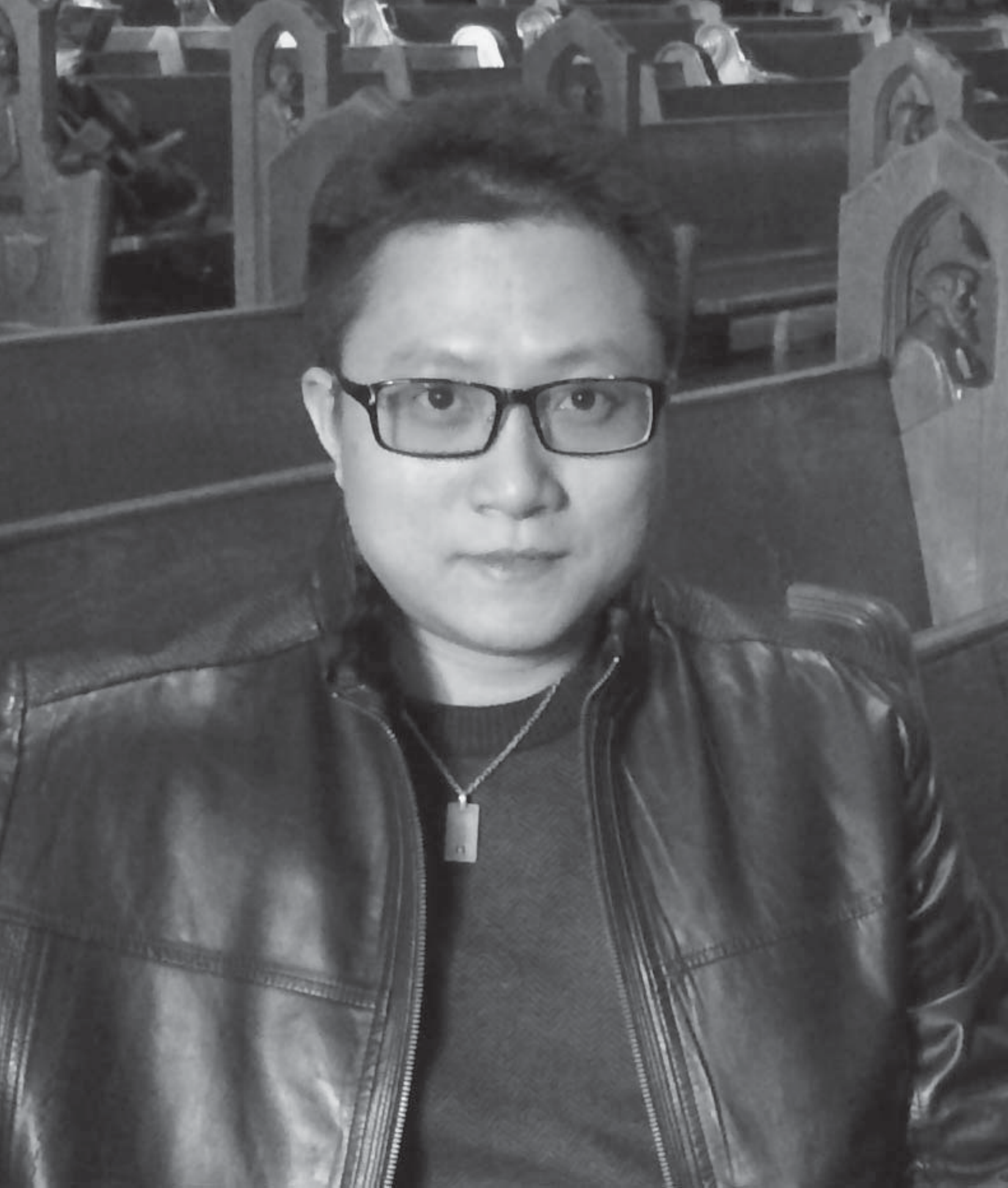}}]{Liang Lin}
	Liang Lin is CEO of DMAI Great China and a full professor of Computer Science in Sun Yat-sen University. He served as the Executive Director of the SenseTime Group from 2016 to 2018, leading the R\&D teams in developing cutting-edge, deliverable solutions in computer vision, data analysis and mining, and intelligent robotic systems.  He has authored or co-authored more than 200 papers in leading academic journals and conferences (e.g., TPAMI/IJCV, CVPR/ICCV/NIPS/ICML/AAAI). He is an associate editor of IEEE Trans, Human-Machine Systems and IET Computer Vision, and he served as the area/session chair for numerous conferences, such as CVPR, ICME, ICCV, ICMR. He was the recipient of Annual Best Paper Award by Pattern Recognition (Elsevier) in 2018, Dimond Award for best paper in IEEE ICME in 2017, ACM NPAR Best Paper Runners-Up Award in 2010, Google Faculty Award in 2012, award for the best student paper in IEEE ICME in 2014, and Hong Kong Scholars Award in 2014. He is a Fellow of IET.
\end{IEEEbiography}
\end{document}